\documentclass[letterpaper, 10 pt, conference]{ieeeconf}
\IEEEoverridecommandlockouts
\usepackage[letterpaper, left=48pt, right=48pt, bottom=50pt, top=57pt]{geometry}
\usepackage{flushend}
\usepackage{lipsum}
\usepackage{graphicx}
\usepackage[center]{caption}
\usepackage{subcaption}
\usepackage{amsmath}
\usepackage{amsfonts}
\usepackage{amssymb}
\usepackage{comment}
\usepackage{bm}
\usepackage{xcolor}
\usepackage{float}
\usepackage{adjustbox}
\usepackage{textcomp}
\usepackage{hyperref}      

\usepackage{cite}
\usepackage[capitalise]{cleveref}

\usepackage{tikz}
\usetikzlibrary{shapes.misc}
\usetikzlibrary{positioning}
\usetikzlibrary{arrows.meta}
\usetikzlibrary{shapes.multipart}
\usetikzlibrary{patterns} 

\newcommand{\R}{\mathbb{R}}

\DeclareMathOperator*{\argmin}{arg\,min}

\makeatletter
\pgfdeclareshape{threedots}{
  \inheritsavedanchors[from=circle]
  \inheritanchorborder[from=circle]
  \inheritanchor[from=circle]{center}
  \inheritanchor[from=circle]{south}
  \inheritanchor[from=circle]{north}
  \inheritanchor[from=circle]{west}
  \inheritanchor[from=circle]{east}
  \backgroundpath{
    \foreach \x in {1,2,3} {
      \pgfmathsetmacro\dotsep{0.3cm}
      \pgfpathcircle{\pgfpoint{\x * \dotsep - 1.3*\dotsep}{0}}{0.05cm}
    }
    \pgfsetfillcolor{black}
    \pgfusepath{fill}
  }
}
\makeatother

\graphicspath{{./figs/}} 

\tikzstyle{startstop} = [rectangle, rounded corners, minimum width=3cm, minimum height=1cm,text centered, draw=black, fill=red!30]
\tikzstyle{process} = [rectangle, minimum width=3cm, minimum height=1cm, text centered, draw=black, fill=orange!30]
\tikzstyle{decision} = [diamond, minimum width=3cm, minimum height=1cm, text centered, draw=black, fill=green!30]

\title{\LARGE \bf Distributed Planning for Rigid Robot Formations with Probabilistic Collision Avoidance}

\author{
    Jeppe Heini Mikkelsen$^{1}$, Vit Kratky$^{2}$, Roberto Galeazzi$^{1}$, Martin Saska$^{2}$, and Matteo Fumagalli$^{1}$ \\
    $^{1}$Technical University of Denmark, Department of Electrical and Photonics Engineering, Denmark\\
    $^{2}$Czech Technical University in Prague, Multi-robot Systems Group, Czech Republic
}

\begin{document}

\maketitle

\begin{abstract}
    This paper presents a distributed method for robots moving in rigid formations while ensuring probabilistic collision avoidance between the robots. The formation is parametrised through the transformation of a base configuration. The robots map their desired velocities into a corresponding desired change in the formation parameters and apply a consensus step to reach agreement on the desired formation and a constraint satisfaction step to ensure collision avoidance within the formation. The constraint set is found such that the probability of collision remains below an upper bound. The method was demonstrated in a manual teleoperation scenario both in simulation and a real-world experiment.
\end{abstract}
\section{Introduction}
In multi-robot systems, coordinated movement in formation is often required for tasks such as inspection and reconnaissance \cite{Ouyang2023FormationReview}. The goal of formation planning algorithms is to compute movements of the robots such that a formation is maintained while they perform their respective tasks and avoid collisions with obstacles and each other. Formation planning algorithms can be categorised as centralised or distributed. In centralised methods, all information is gathered at a central location where the plans are computed, while in distributed methods all robots participate in the computation of the plans and coordinate via communication, requiring inter-robot coordination mechanisms \cite{Ouyang2023FormationReview}. Centralised methods can be simpler to implement since they do not require inter-robot coordination mechanisms, but have the drawback of system dependency on a single computer representing a single point of failure. Furthermore, formation planning algorithms can be optimal or feasible. Optimal methods aim to minimise a cost function while adhering to constraints and feasible methods aim to find sub-optimal movements that adhere to constraints. Feasible methods, although sub-optimal, are often less computationally demanding and thus more suitable for robots with fast dynamics or limited computational capabilities \cite{Ouyang2023FormationReview}. Lastly, formations can be rigid, where the formation maintains a fixed geometric shape, or non-rigid, where the formation is allowed to deform. In this paper a feasible distributed planning algorithm for rigid formations is presented.

Survey papers on formation planning and control algorithms can be found in \cite{Chen2005FormationConsideration,Oh2015AControl,Ouyang2023FormationReview}. A rigid formation can be treated as a virtual rigid body (VRB), where the aim of the formation planning algorithm is to compute the desired states of the VRB according to mission specifications. Schneider et al. \cite{Schneider2003ANavigation} introduce an artificial potential field (APF) \cite{Khatib1985Real-timeRobots} for maintaining a formation and moving it to a target state while avoiding collisions. In \cite{Lawton2003AManeuvers}, a linear controller is used to move the robots towards a desired formation and consensus is used to reach agreement on the translation of the formation. In \cite{RenConsensusFormations}, consensus for a second-order system is used to reach agreement on the translation of a formation. In \cite{Ren2008DistributedControl}, consensus is used by the robots to estimate the desired motion of the VRB, which is known to a subset of the robots, and also used to track the VRB. In \cite{Yoshioka2008FormationStructure}, a consensus strategy for reaching agreement on the state of a VRB, with a mapping from the VRB state derivative to the velocities of the robots, was presented. In \cite{Ayanian2010AbstractionsObstacles}, the VRB is used to define a cell containing the robots where the robots are free to move, and the VRB translates, rotates, and scales to move in the environment while avoiding collisions. In \cite{Zhou2018AgileStructures}, an APF is applied both to the VRB and the robots, where the potential acting on the VRB is calculated at a central node and the potentials acting on the robots are calculated locally on each robot. In \cite{Derenick2007ConvexFormations}, convex optimisation is used to find the optimal similarity transformation of a configuration that minimises the distance travelled by the robots, with constraints on the velocities of the robots and environmental constraints on the formation. In \cite{Alonso-Mora2015Multi-robotProgramming,Alonso-Mora2016DistributedConsensus}, a method for navigating a formation toward a desired state, by iteratively computing the optimal VRB within the largest convex polytope containing the current formation, is presented. In \cite{Mosteo2017OptimalFormations}, the authors demonstrate that finding the optimal rotation, translation, and assignment of the robots in a VRB can be solved separately. In \cite{Wang2023Collision-FreeTopologies}, consensus is used to reach agreement on the velocities of the robots and the displacement of the formation, and an APF is used to achieve collision avoidance. Instead of encoding the formation through the similarity transformation of a VRB, it can be encoded through relative distance constraints between the robots. In \cite{Cortes2009GlobalNetworks}, the formation is maintained by moving the robots according to the gradient of a stress function between the robots. In \cite{Lei2015ConsensusControl} the authors propose a bounded consensus algorithm to reach agreement on the heading of the formation while constraining the velocities of the robots, with distance constraints on the relative positions of the robots to maintain formation. In \cite{Wu2020AUAVs}, consensus is used to align the velocities of the robots and a control term is added to drive them towards positions where a desired relative distance is maintained. Collisions are avoided by changing their heights, and obstacle avoidance is achieved using particle swarm optimisation.

The primary contribution of this paper is the development of a distributed method for controlling rigid robot formations with a probabilistic approach to collision avoidance, which has not yet been addressed in the literature to the best of our knowledge. The proposed method maps the local desired velocities of the robots into global formation parameters, while ensuring that the probability of collision remains below a predefined upper bound. Unlike traditional methods that primarily utilise artificial potential fields and often neglect the inherent uncertainties in robot positioning, the presented method addresses these uncertainties by dynamically adjusting the constraints on the transformation of the formation's base configuration. The proposed approach is validated through both simulation and real-world experiments, demonstrating its effectiveness in maintaining formation integrity while ensuring safe operation in cluttered environments. The code used in the simulation and experiment results can be found at \url{github.com/JeppHMikk/consensus-formation-flying}.

Throughout the paper, we adopt the following notation style: italic symbols $x/X$ denote scalars; bold italic symbols \(\bm{x}/\bm{X}\) denote vectors; bold non-italic symbols \(\mathbf{x}/\mathbf{X}\) denote matrices; and calligraphic symbols \(\mathcal{X}\) denote sets. Notations \(\R_{\geq0}\) and \(\R_{>0}\) denote non-negative and strictly positive real numbers, respectively.

\section{Problem Description and Approach}\label{sec:problem}
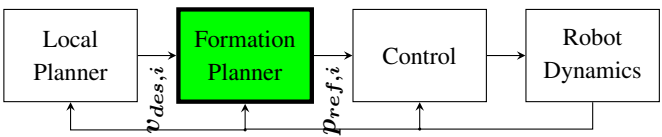
\begin{figure}[ht]
    \centering
    \resizebox{\linewidth}{!}{
    \begin{tikzpicture}[scale=0.35]
        \node [draw,
            fill=white,
            minimum width=1.5cm,
            minimum height=1.2cm,
            text width=1.5cm,
            align=center
        ]  (local_planner) at (0,0) {\small{Local Planner}};

        \node [draw,
            fill=green,
            fill opacity=0.5,
            text opacity=1,
            minimum width=1.5cm,
            minimum height=1.2cm,
            text width=1.5cm,
            align=center,
            right = 0.5cm of local_planner,
            ultra thick
        ] (formation_planner) {\small{Formation Planner}};

        \node [draw,
            fill=white,
            minimum width=1.5cm,
            minimum height=1.2cm,
            text width=1.5cm,
            align=center,
            right = 0.5cm of formation_planner
        ] (control) {\small{Control}};

        \node [draw,
            fill=white,
            minimum width=1.5cm,
            minimum height=1.2cm,
            text width=1.5cm,
            align=center,
            right = 0.5cm of control
        ] (robot) {\small{Robot Dynamics}};

        \draw[-stealth] (local_planner.east) -- (formation_planner.west) node[midway,rotate=90,left]{$\bm{v_{des,i}}$};

        \draw[-stealth] (formation_planner.east) -- (control.west) node[midway,rotate=90,left]{$\bm{p_{ref,i}}$};

        \draw[-stealth] (control.east) -- (robot.west);

        \draw[-stealth] (robot.south) -- ++ (0,-1.0) -| (control.south);

        \draw[-stealth] (control.south) ++ (0,-1.0) -| (formation_planner.south);

        \draw[-stealth] (formation_planner.south) ++ (0,-0.97) -| (local_planner.south);

        \filldraw[black] (control.south) ++ (0,-1.0) circle (1pt);

        \filldraw[black] (formation_planner.south) ++ (0,-0.97) circle (1pt);
        
    \end{tikzpicture}}
    \caption{System architecture: Each robot has a local planner, a formation planner and a control system \cite{Mikkelsen2023DistributedConfiguration}.}
    \label{fig:sys_architecture}
\end{figure}

Consider a swarm of $N$ planar robots with indices $\mathcal{V} = \{1,\dots,N\}$, where the position of robot $i$ at time $t$ is denoted by $\bm{p_i}(t) \in \R^2$. 
This paper presents a distributed algorithm for ensuring that the robots fly in a rigid formation and avoid collisions, while each robot in the formation attempt to track a local desired velocity $\bm{v_{des,i}}$, supplied by local planners with higher level goals. An example of a local planner for manual teleoperation of a formation is presented in \cref{sec:application}. 
It is assumed that the robots have onboard controllers capable of tracking a reference position $\bm{p_{ref,i}}$. Furthermore, the physical space that each robot takes up is represented as a minimum circumscribing circle with radius $r_i$ and centre at position $\bm{p_i}(t)$. Each robot has a Gaussian distributed position estimate
\begin{equation}
    \bm{\hat{p}_i}(t) \sim \mathcal{N}(\bm{\bar{p}_i}(t),\mathbf{\Sigma_i})
\end{equation}
with mean $\bm{\bar{p}_i}(t)\in\R^2$ and covariance matrix $\mathbf{\Sigma_i}\in\R^{2\times2}$. Since the algorithm will have to rely on the position estimate of the robots, it also has to ensure that the probability of collisions within the formation remain below some upper bound $\Bar{p}_{coll}\in\R_{>0}$. All of this is achieved by injecting an intermediary formation planner between the local planner and controller on each robot, where the desired velocities are supplied by local planners; see \cref{fig:sys_architecture}.

\section{Virtual Rigid Body Transformation}\label{sec:transformation}
Like in \cite{Mikkelsen2023DistributedConfiguration}, the position of the robots in the formation is represented as a virtual rigid body (VRB), which is described through the transformation of a base configuration $\mathcal{B}=\{\bm{c_1},\dots,\bm{c_N}\}$, where $\bm{c_i}\in\R^2$ is the position in the base configuration associated with robot $i$. The position $\bm{q_i}\in\R^2$ of robot $i$ in the VRB is found as the scaling, rotation, and translation of $\bm{c_i}$
\begin{equation}\label{eq:transformation}
    \bm{q_i} = \mathbf{R}\mathbf{S}\bm{c_i} + \bm{t} \ \forall \ i \in \mathcal{V},
\end{equation}
where $\mathbf{S}\in\R_{>0}^{2\times2}$ is a diagonal scaling matrix, $\mathbf{R}\in SO(2)$ is a rotation matrix, and $\bm{t}\in\R^2$ is a translation vector; see \cref{fig:transformation}
\begin{equation}\label{eq:transformation_components}
    \mathbf{R} = 
    \begin{bmatrix}
        \cos\varphi & -\sin\varphi \\
        \sin\varphi &  \cos\varphi
    \end{bmatrix}, \
    \mathbf{S} =
    \begin{bmatrix}
        s_x & 0 \\
        0   & s_y
    \end{bmatrix}, \
    \bm{t} =
    \begin{bmatrix}
        t_x \\
        t_y
    \end{bmatrix}.
\end{equation}
The parameter vector of the transformation is denoted as
\begin{equation}\label{eq:param}
    \begin{gathered}
    \bm{\eta} = (\varphi,\bm{s},\bm{t}) \in \R^5, \\ \quad \bm{s} = (s_x,s_y)\in\R^2, \quad \bm{t} = (t_x,t_y)\in\R^2.
    \end{gathered}
\end{equation}
\tikzset{cross/.style={cross out, draw=black, fill=none, minimum size=2*(#1-\pgflinewidth), inner sep=0pt, outer sep=0pt}, cross/.default={5pt}}
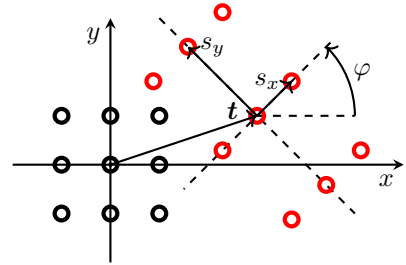
\begin{figure}[ht]
    \centering
    \begin{tikzpicture}[scale=0.65]

        \draw[thick,-stealth] (-2,0) -- (6,0) node[anchor=north east]{$x$};
        \draw[thick,-stealth] (0,-2) -- (0,3) node[anchor=north east]{$y$};
        \filldraw[black] (0,0) circle (1pt);        
        \draw[ultra thick] (-1,-1) circle (4pt);
        \draw[ultra thick] (-1,0) circle (4pt);
        \draw[ultra thick] (-1,1) circle (4pt);
        \draw[ultra thick] (0,-1) circle (4pt);
        \draw[ultra thick] (0,0) circle (4pt);
        \draw[ultra thick] (0,1) circle (4pt);
        \draw[ultra thick] (1,-1) circle (4pt);
        \draw[ultra thick] (1,0) circle (4pt);
        \draw[ultra thick] (1,1) circle (4pt);
        
        \draw[thick,dashed] (1,3) -- (5,-1);
        \draw[thick,dashed] (4.5,2.5) -- (1.5,-0.5);
        \draw[thick,dashed] (3,1) -- (5,1);


        \draw[ultra thick,red] (3.7071,-1.1213) circle (4pt);
        \draw[ultra thick,red] (2.2929,0.2929) circle (4pt);
        \draw[ultra thick,red] (0.8787,1.7071) circle (4pt);
        \draw[ultra thick,red] (4.4142,-0.4142) circle (4pt);
        \draw[ultra thick,red] (3.0000,1.0000) circle (4pt);
        \draw[ultra thick,red] (1.5858,2.4142) circle (4pt);
        \draw[ultra thick,red] (5.1213,0.2929) circle (4pt);
        \draw[ultra thick,red] (3.7071,1.7071) circle (4pt);
        \draw[ultra thick,red] (2.2929,3.1213) circle (4pt);

        \draw[thick,->] (5,1) arc (0:45:2);
        \node at (5.15,1.9) {$\varphi$};
        \draw[thick,->] (0,0) -- (3,1);
        \node at (2.5,1.1) {$\bm{t}$};
        \draw[thick,->] (3,1) -- (3.71,1.71);
        \node at (3.21,1.71) {$s_x$};
        \draw[thick,->] (3,1) -- (1.59,2.41);
        \node at (2.09,2.41) {$s_y$};

    \end{tikzpicture}
    \caption{Transformation of a unit grid swarm configuration with $\varphi=\pi/4$, $s_x=1$, $s_y=2$, $t_x=3$, and $t_y=1$. \textbf{Black circle}: base configuration. \textcolor{red}{\textbf{Red circle}}: transformed base configuration \cite{Mikkelsen2023DistributedConfiguration}.}
    \label{fig:transformation}
\end{figure}

The base configuration $\mathcal{B}$ is specified by the operator and can have any desired shape. It is assumed that the centroid of the configuration is at the origin, i.e.,
\begin{equation}
    \sum_{i\in\mathcal{V}} \bm{c_i} = \bm{0_{2\times1}},
\end{equation}
which can be achieved for any configuration by shifting each coordinate by the centroid of the formation, i.e.
\begin{equation}
    \bm{c_i} \leftarrow \bm{c_i} - \frac{1}{N}\sum_{j\in\mathcal{V}} \bm{c_j} \ \forall \ i \in \mathcal{V},
\end{equation}
ensuring that the formation rotates around its centroid.
\section{Method}
The method for ensuring formation flying presented in this paper is based on the work in \cite{Mikkelsen2023DistributedConfiguration} with an extension to handle probabilistic collision avoidance. Instead of acting on the robots directly, the method is applied to a VRB, which the robots then attempt to track using onboard position controllers, ensuring that the dynamics of the robots is decoupled from the planning of the formation. The method consists of five steps: \textit{tracking}, \textit{consensus}, \textit{constraint satisfaction}, \textit{reference generation}, and \textit{parameter update}, where the constraint satisfaction step have been extended to handle probabilistic collision avoidance. Each robot has a local parameter $\bm{\eta_i}(t)\in\R^5 \ \forall \ i\in\mathcal{V}$ and position reference $\bm{p_{ref,i}}(t)\in\R^2 \ \forall \ i\in\mathcal{V}$, that are updated according to the five steps.
\subsection{Tracking}
Each robot in the VRB has a desired velocity $\bm{v_{des,i}}\in\R^2$, which is found according to some higher level planner. In the tracking step, the time-derivative of the local parameters, which ensures that the robot tracks its desired velocity, is found as\begin{equation}\label{eq:step1}
    \frac{d}{dt}\bm{\eta_i}(t) = \mathbf{J_i^+}\bm{v_{des,i}}(t),
\end{equation}
where $\mathbf{J_i}$ is the Jacobian of the transformation in \ref{eq:transformation} with respect to $\bm{\eta_i}(t)$
\begin{multline}\label{eq:jacobian}
    \mathbf{J_i} = \left[
    \begin{matrix}
        -\sin\varphi_i s_{x,i} c_{x,i} - \cos\varphi_i s_{y,i} c_{y,i} \\
        \cos\varphi_i s_{x,i} c_{x,i} -\sin\varphi_i s_{y,i} c_{y,i}
    \end{matrix} \right. \dots\\
    \left.
    \begin{matrix}
        \cos\varphi_i c_{x,i} & -\sin\varphi_i c_{y,i} & 1 & 0 \\
        \sin\varphi_i c_{x,i} & \cos\varphi_i c_{y,i} & 0 & 1
    \end{matrix} \right],
\end{multline}
and $(\cdot)^+$ denotes the right Moore-Penrose pseudo-inverse computed as
\begin{equation}
    \mathbf{J_i^+} = \mathbf{J_i^\top}(\mathbf{J_i}\mathbf{J_i}^\top)^{-1}.
\end{equation}
Since the individual desired velocities of the robots may not conform to a feasible formation motion, the tracking step in \eqref{eq:step1} may not produce the same parameter derivative across the robots, leading to discrepancies in the parameters over time and bringing the robots out of formation.
\subsection{Consensus}
To ensure that the robots remain in formation, a consensus step is applied to \ref{eq:step1}
\begin{equation}
        \frac{d}{dt}\bm{\eta_i}(t) = \mathbf{J_i^+}\bm{v_{des,i}}(t)\underbrace{-\lambda_i\sum_{j\in\mathcal{V}\setminus\{i\}} (\bm{\eta_i}(t) - \bm{\eta_j}(t))}_{\textbf{consensus step}},
\end{equation}
where $\lambda_i \in \R_{>0}$ determines the stiffness of the formation.
\subsection{Constraint Satisfaction}
\begin{figure}[ht]
    \centering
    \begin{tikzpicture}[scale=0.75]
        \draw[draw=red, fill=red!10] (0,0) circle (1.5) node[red,opacity=1,anchor=north east]{$\mathcal{B}_R$};
        \draw[red,-stealth] (0,0) -- (1.061,-1.061) node[red,pos=0.5,anchor=west]{$R$};
        \draw[blue] (0.7442,1.3024) -- ++(-2,1.1429);
        \draw[blue] (0.7442,1.3024) -- ++(2,-1.1429);
        \fill[pattern=dots, pattern color=blue,rotate=60.2551] (-2,-2.3035) rectangle (1.5,2.3035);
        \node[blue] at (0.75,-2) {$\mathcal{H}$};
        \draw[-stealth] (-2,0) -- (3,0) node[anchor=north east]{$\delta x$};
        \draw[-stealth] (0,-2) -- (0,3) node[anchor=north east]{$\delta y$};
        \filldraw[color=black, fill=black](0,0) circle (0.05);
        \draw[xshift=1cm,yshift=1.75cm,rotate=-100.6806] (0,0) ellipse (1.0253cm and 1.5cm);
        \draw[-stealth] (0,0) -- (1,1.75) node[pos=0.25,anchor=west]{$\bm{\bar{\delta}_{ij}}$} node[pos=1.0,anchor=west]{$\mathbf{\Sigma_{ij}}$};
    \end{tikzpicture}    
    \caption{Distance vector covariance ellipse, ball region and hyperplane approximation.}
    \label{fig:hyperplane_approx}
\end{figure}
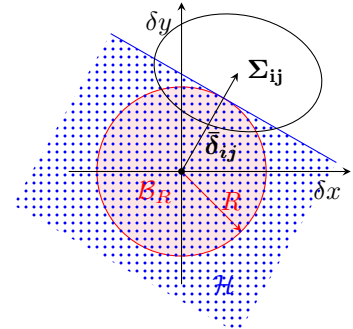

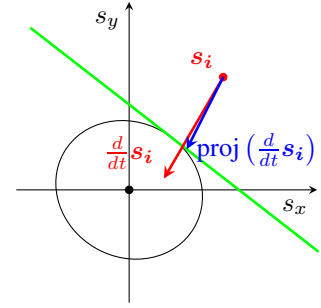
\begin{figure}
    \centering
    \begin{tikzpicture}
        \draw[-stealth] (-1.5,0) -- (2.5,0) node[anchor=north east]{$s_x$};
        \draw[-stealth] (0,-1.5) -- (0,2.5) node[anchor=north east]{$s_y$};
        \filldraw[color=black, fill=black](0,0) circle (0.05);
        \draw[rotate=150.0004] (0,0) ellipse (1cm and 0.8944cm);
        \filldraw[color=red, fill=red](1.25,1.5) circle (0.05) node[anchor=south east]{$\bm{s_i}$};
        \draw[-,color=green,line width=1pt] (-1.3145,2.1573) -- (2.5186,-0.8237);
        \draw[-stealth,color=red,line width=1pt] (1.25,1.5) -- (0.4625,0.15) node[anchor=south east]{$\frac{d}{dt}\bm{s_i}$};
        \draw[-stealth,color=blue,line width=1pt] (1.25,1.5) -- (0.7655,0.5397) node[anchor= west]{$\text{proj}\left(\frac{d}{dt}\bm{s_i}\right)$};
    \end{tikzpicture}    
    \caption{Robots $i$ scaling parameter derivative $\frac{d}{dt}\bm{s_i}$ (red vector) is projected onto a linear approximation (green line) of a quadratic constraint (black ellipse).}
    \label{fig:quad_const}
\end{figure}

The two prior steps ensure that the robots attempt to track their desired velocities while agreeing on the formation. However, this might result in infeasible formations that lead to collisions between robots. To ensure that the formation is feasible, a constraint satisfaction step has to be applied. In \cite{Mikkelsen2023DistributedConfiguration} this was handled by constraining the scaling parameters of the formation to remain within a static constraint set. This works when the exact positions of the robots are known or if the covariance is constant. However, most robot systems rely on some sort of estimation where the covariance of the position estimate might be non-constant. Therefore, the constraint set must be dynamic such that if one or more of the robots become increasingly uncertain about its position, the constraints will change to ensure that the probability of collision remains below an upper bound $\bar{p}_{coll}$. A collision is assumed to be inevitable if the robots are within some distance $\varepsilon\in\R_{\geq 0}$ from each other. Since the robots are assumed to be spherical, the condition for a collision between robot $i$ and robot $j$ can be expressed as
\begin{equation}
    collision(\bm{p_i},\bm{p_j}) = True \Leftrightarrow ||\bm{p_j} - \bm{p_i}||_2 \leq r_i + r_j + \varepsilon,
\end{equation}
where $r_i$ and $r_j$ are the radii of the robots. Since the position estimates of the robots follow Gaussian distributions, the distance vector between them also follow a Gaussian distribution
\begin{equation}
    \bm{\hat{\delta}_{ij}} \sim \mathcal{N}(\bm{\bar{\delta}_{ij}}, \mathbf{\Sigma_{ij}})
\end{equation}
where $\bm{\bar{\delta}_{ij}} = \bm{\bar{p}_j} - \bm{\bar{p}_i}$ and $\mathbf{\Sigma_{ij}} = \mathbf{\Sigma_{i}} + \mathbf{\Sigma_{j}}$. The probability of a collision between robot $i$ and $j$ is therefore found by integrating the distance vector likelihood over a ball $\mathcal{B}_R$ with radius $R=r_i + r_j + \varepsilon$ centred at the origin of the distance vector
\begin{equation}
    p_{coll}(\bm{p_i},\bm{p_j}) = \int_{\mathcal{B}_R} \text{Pr}(\bm{\delta_{ij}}|\bm{\hat{\delta}_{ij}}, \mathbf{\Sigma_{ij}}) \ d\bm{\delta_{ij}},
\end{equation}
where $\text{Pr}(\bm{\delta_{ij}}|\bm{\bar{\delta}_{ij}}, \mathbf{\Sigma_{ij}})$ is the PDF of the Gaussian evaluated at $\bm{\delta_{ij}}$. The integral does not have a closed form solution and can therefore not easily be used in a formation planning algorithm without resorting to numerical integration, which is deemed too computationally demanding. Graphically, the problem of integrating over the ball $\mathcal{B}_R$ can be seen in \cref{fig:hyperplane_approx}.
Instead of integrating over the ball region, the PDF can be integrated over a hyperplane $\mathcal{H} = \{\delta_{ij} \ | \ \bm{n_{ij}^\top}\delta_{ij} \leq d_{ij}\}$ containing the ball region
\begin{equation}
    \hat{p}_{coll}(\bm{p_i},\bm{p_j}) = \int_{\mathcal{H}} \text{Pr}(\bm{\delta_{ij}}|\bm{\bar{\delta}_{ij}}, \mathbf{\Sigma_{ij}}) \ d\bm{\delta_{ij}}.
\end{equation}
Since the ball region is contained within the hyperplane, the integral over the hyperplane upper bounds the integral over the ball region
\begin{equation}
    \hat{p}_{coll}(\bm{p_i},\bm{p_j}) \geq p_{coll}(\bm{p_i},\bm{p_j}),
\end{equation}
and therefore by ensuring that $\hat{p}_{coll}(\bm{p_i},\bm{p_j}) \leq \bar{p}_{coll}$, then $p_{coll}(\bm{p_i},\bm{p_j}) \leq \bar{p}_{coll}$ \cite{Zhu2019Chance-ConstrainedEnvironments}. The hyperplane has normal vector $\bm{n_{ij}} = \bm{\bar{\delta}_{ij}}(||\bm{\bar{\delta}_{ij}}||_2)^{-1}$ and bias $d_{ij} = r_i + r_j + \varepsilon$. Projecting the distribution onto the normal vector
\begin{equation}
    \rho_{ij} \sim \mathcal{N}(\hat{\rho}_{ij},\sigma_{ij}^2)
\end{equation}
where $\hat{\rho}_{ij} = \bm{n_{ij}^\top}\bm{\bar{\delta}_{ij}}$ and $\sigma_{ij}^2 = \bm{n_{ij}^\top}\mathbf{\Sigma_{ij}}\bm{n_{ij}}$, the probability of collision can be calculated as
\begin{equation}
    \hat{p}_{coll}(\bm{p_i},\bm{p_j}) = \int_{-\infty}^{d_{ij}} \text{Pr}(\rho_{ij}|\hat{\rho}_{ij}, \sigma_{ij}^2) \ d\rho_{ij}.
\end{equation}
This integral does not have a closed-form solution either. However, since $\rho_{ij}$ follows a uni-variate Gaussian distribution, the relation between the confidence interval ($CI$) and the standard deviation $\sigma_{ij}$ is known. By ensuring that 
\begin{equation}
    \hat{\rho}_{ij} \geq r_i + r_j + \varepsilon + \xi \sigma_{ij},
\end{equation}
it can be guaranteed that $\hat{p}_{coll} \leq \bar{p}_{coll}$, where $\xi$ is the relation between the standard deviation $\sigma_{ij}$ and the confidence interval $CI = 1 - 2\bar{p}_{coll}$, e.g., for $\bar{p}_{coll} = 1.5e-3$ the confidence interval is $CI=0.997$ which corresponds to $\xi=3$. By substituting the mean and covariance with the projection, the inequality can be rewritten as
\begin{equation}
    \bm{n_{ij}^\top}\bm{\bar{\delta}_{ij}} \geq r_i + r_j + \varepsilon + \xi\left(\bm{n_{ij}^\top}\mathbf{\Sigma_{ij}}\bm{n_{ij}}\right)^{1/2},
\end{equation}
which when inserting $\bm{n_{ij}} = \bm{\bar{\delta}_{ij}}(||\bm{\bar{\delta}_{ij}}||_2)^{-1}$ becomes
\begin{equation}
    ||\bm{\bar{\delta}_{ij}}||_2 \geq r_i + r_j + \varepsilon + \xi\left(\frac{\bm{\bar{\delta}_{ij}^\top}\Sigma_{ij}\bm{\bar{\delta}_{ij}}}{\bm{\bar{\delta}_{ij}^\top}\bm{\bar{\delta}_{ij}}} \right)^{1/2}.
\end{equation}
The fourth term on the right-hand side contains the Rayleigh quotient which is upper bounded by the largest eigenvalue $\lambda_{ij,\max}$ of $\mathbf{\Sigma_{ij}}$, simplifying the inequality to
\begin{equation}\label{eq:ineq_dist}
    ||\bm{\bar{\delta}_{ij}}||_2 \geq r_i + r_j + \varepsilon + \xi\sqrt{\lambda_{ij,\max}}.
\end{equation}
Since the left-hand and right-hand side of the inequality are guaranteed to be positive, both can be squared while preserving the inequality
\begin{equation}
    \bm{\bar{\delta}_{ij}^\top}\bm{\bar{\delta}_{ij}} \geq \left(r_i + r_j + \varepsilon + \xi\sqrt{\lambda_{ij,\max}}\right)^2.
\end{equation}
Assuming that robot $i$ and $j$ have reached consensus on the transformation or that the difference is negligible, the mean distance vector can be found using the transformation in \ref{eq:transformation} as
\begin{equation}
    \bm{\bar{\delta}_{ij}} = \mathbf{R}\mathbf{\Delta C_{ij}}\bm{s_i}
\end{equation}
where $\mathbf{\Delta C_{ij}} = \text{diag}(\bm{c_j} - \bm{c_i})$. This can then be inserted into the inequality
\begin{equation}
    \bm{s_i^\top}\mathbf{\Delta C_{ij}^\top}\mathbf{R^\top}\mathbf{R}\mathbf{\Delta C_{ij}}\bm{s_i} \geq \left(r_i + r_j + \varepsilon + \xi\sqrt{\lambda_{ij,\max}}\right)^2.
\end{equation}
Since the rotation matrix is an orthogonal matrix, i.e., $\mathbf{R^{-1}} = \mathbf{R^\top}$, this simplifies to
\begin{gather}
    \bm{s_i^\top}\mathbf{\Gamma_{ij}}\bm{s_i} \geq \gamma_{ij}, \\
    \mathbf{\Gamma_{ij}} = \mathbf{\Delta C_{ij}^\top}\mathbf{\Delta C_{ij}}, \\ 
    \gamma_{ij} = \left(r_i + r_j + \varepsilon + \xi\sqrt{\lambda_{ij,\max}}\right)^2.
\end{gather}
Thus, by substituting the Rayleigh quotient with the maximum eigenvalue $\lambda_{ij,\max}$, the inequality has become rotation invariant. This gives a non-convex quadratic inequality constraint on the scaling parameters for each robot pair, and the total constraint set becomes the intersection of all the quadratic inequality constraint sets. This set is non-convex, and consensus requires that the constraint sets are convex to ensure convergence. Therefore, the quadratic constraints are linearised, see 
\cref{fig:quad_const}.
The quadratic inequality constraint is approximated as a linear inequality constraint
\begin{equation}
    \bm{a_{ij}^\top}\bm{s_i} \geq b_{ij},
\end{equation}
by linearising it around the current scaling parameter $\bm{s_i}(t)$ according to
\begin{gather}
    \bm{a_{ij}} = \mathbf{\Gamma_{ij}}\bm{s_i}(t) \\ \quad b_{ij} = \bm{s_i^\top}(t)\mathbf{\Gamma_{ij}}\bm{s_i}(t) - \alpha\bm{s_i^\top}(t)\mathbf{\Gamma_{ij}}\mathbf{\Gamma_{ij}}\bm{s_i}(t),
\end{gather}
where $\alpha$ is found as
\begin{gather}
    \alpha = \min\left(\frac{-b \pm \sqrt{b^2 - 4ac}}{2a}\right), \\
    a = \bm{s_i^\top}(t)\mathbf{\Gamma_{ij}^\top\Gamma_{ij}\Gamma_{ij}}\bm{s_i}(t), \\
    b = -2\bm{s_i^\top}(t)\mathbf{\Gamma_{ij}^\top\Gamma_{ij}}\bm{s_i}(t), \\
    c = \bm{s_i^\top}(t)\mathbf{\Gamma_{ij}}\bm{s_i}(t) - \gamma_{ij}.
\end{gather}
Using this approach, a linear inequality constraint on the scaling parameter vector is generated for every other robot and concatenated together into a single inequality constraint
\begin{equation}
    \mathbf{A_i}\bm{s_i} \geq \bm{b_i}.
\end{equation}
To ensure that the inequality constraint is being met at all times, the parameter derivative of robot $i$ at time $t$ is projected onto the constraint set
\begin{equation}\label{eq:proj_step}
    \begin{split}
        \frac{d}{dt}\bm{\eta_i}(t) = \text{proj}(\mathbf{J_i^+}\bm{v_{des,i}}(t) \dots \\ - \lambda_i\sum_{j\in\mathcal{V}\setminus\{i\}} (\bm{\eta_i}(t) - \bm{\eta_j}(t)))
    \end{split}
\end{equation}
according to
\begin{equation}
    \text{proj}\left(\frac{d}{dt}\bm{\eta_i}\right) = \left(\frac{d}{dt}\varphi,\text{proj}\left(\frac{d}{dt}\bm{s_i}\right),\frac{d}{dt}\bm{t_i}\right),
\end{equation}
where
\begin{equation}
    \begin{aligned}
        \text{proj}\left(\frac{d}{dt}\bm{s_i}\right) &= \argmin_{\bm{\Delta s_i}} \frac{1}{2}||\frac{d}{dt}\bm{s_i}(t) - \bm{\Delta s_i}||_2^2, \\
        s.t. & \quad \mathbf{A_i}(\bm{s_i}(t) + \bm{\Delta s_i}) \geq \bm{b_i}.
    \end{aligned}
\end{equation}
The projection of the scaling parameter derivative is cast as a linear inequality constrained quadratic program and is solved using a custom implementation of the active-set algorithm. By ensuring that the scaling parameter derivative vector is never allowed to leave the constraint set, the scaling parameter is guaranteed to remain within it for all time. 

The parameter derivative also needs to be constrained to ensure that the velocities of the robots in the VRB are not faster than what the actual robots can track. Therefore, the parameter derivative is scaled by
\begin{equation}
    \begin{split}
        \frac{d}{dt}\bm{\eta_i}(t) \leftarrow \beta\frac{d}{dt}\bm{\eta_i}(t),
    \end{split}
\end{equation}
where $\beta$ is found according to
\begin{equation}
    \beta = 
    \begin{cases}
        v_{\max}(||\mathbf{J_i}\frac{d}{dt}\bm{\eta_i}||_2)^{-1} \quad if \quad ||\mathbf{J_i}\frac{d}{dt}\bm{\eta_i}||_2 \geq v_{\max}, \\
        1 \quad else,
    \end{cases}
\end{equation}
with $v_{\max}$ being the maximum allowed velocity of the robots.
\subsection{Reference Generation}
The current parameters for each robot is used to calculate the position references in the VRB using \ref{eq:transformation}
\begin{equation}
    \bm{p_{ref,i}}(t) = \mathbf{R_i}\mathbf{S_i}\bm{c_i} + \bm{t_i},
\end{equation}
where the current parameter vector $\bm{\eta_i}(t)$ of robot $i$ is used to compute the rotation matrix $\mathbf{R_i}$, scaling matrix $\mathbf{S_i}$, and translation vector $\bm{t_i}$. The position reference is then supplied to the robots onboard controller.
\subsection{Parameter Update}
The algorithm runs in discrete time steps of size $\Delta t$. Hence, the parameter vector is updated using Eulers method
\begin{equation}
    \bm{\eta_i}(t+\Delta t) = \bm{\eta_i}(t) + \Delta t \frac{d}{dt}\bm{\eta_i}(t).
\end{equation}
To ensure that the projection step in \ref{eq:proj_step} is valid $\Delta t \leq 1$.

\section{Application}\label{sec:application}
In this paper, the presented method is used for manual teleoperation of drone formations in cluttered environments. The desired velocities of the drones in the VRB is a combination of a commanded change in the formation from an operator, and a velocity for ensuring collision avoidance with obstacles which is computed locally
\begin{equation}
    \bm{v_{des,i}}(t) = \mathbf{J_i}\frac{d}{dt}\bm{\eta_{des}}(t) + \bm{v_{rep,i}}(t).
\end{equation}
This allows an operator to control a number of drones, which is not possible to do if they are all operated individually, while the drones autonomously ensure that the formation changes such that the robots do not collide with obstacles in their vicinity and each other. Thereby, even if the operator loses connection with the formation, the robots are still able to maintain the formation and avoid collisions until connection is recovered or an automatic fail-safe procedure is performed. To avoid collisions with obstacles $\bm{v_{rep,i}}(t)$ is computed using a repulsive APF. The positions $\mathcal{O}\subseteq\R^2$ that the obstacles inhabit in the environment generate a potential that grows as the VRB gets closer to the obstacle boundary
\begin{equation}
    U_{rep,i} = \begin{cases}
        \frac{1}{2}\psi\left(\frac{1}{\rho_i} - \frac{1}{\rho_0}\right)^2, \ \text{if} \ \rho \leq \rho_0, \\
        0, \ \text{else}.
    \end{cases}
\end{equation}
where $\psi$ is the strength of the repulsive field, $\rho_0$ is the distance at which the potential starts growing and $\rho_i$ is the distance from the current desired position of robot $i$ to the closest obstacle minus the desired clearance
\begin{equation}
    \rho_i = \min_{\bm{p_{o}}\in\mathcal{O}} ||(\mathbf{R_iS_i}\bm{c_i} + \bm{t_i}) - \bm{p_{o}}||_2 - (\varepsilon + r_i +\xi\sqrt{\lambda_{i,\max}}),
\end{equation}
where the radius of the robot is enlarged by $\xi$ times the largest standard deviation $\sqrt{\lambda_{i,\max}}$ of $\mathbf{\Sigma_i}$.
By having the robots follow the negative gradient of the potential field, they are pushed away from the obstacles
\begin{equation}
    \begin{aligned}
        \bm{v_{rep,i}}(t) &= -\nabla U_{rep,i} \\
        &= \psi\left(\frac{1}{\rho(\bm{p_i})} - \frac{1}{\rho_0}\right)\frac{1}{\rho(\bm{p_i})^2}\nabla\rho(\bm{p_i}),
    \end{aligned}
\end{equation}
where
\begin{equation}
    \nabla\rho(\bm{p_i}) = \frac{\bm{p_i} - \bm{p_o}}{||\bm{p_i} - \bm{p_o}||_2}.
\end{equation}
Due to the consensus step, this induces a change in the formation such that collision avoidance is achieved while the robots remain in formation.
\section{Simulation Results}
\begin{figure}[ht]
    \centering
    \includegraphics[width=.75\linewidth]{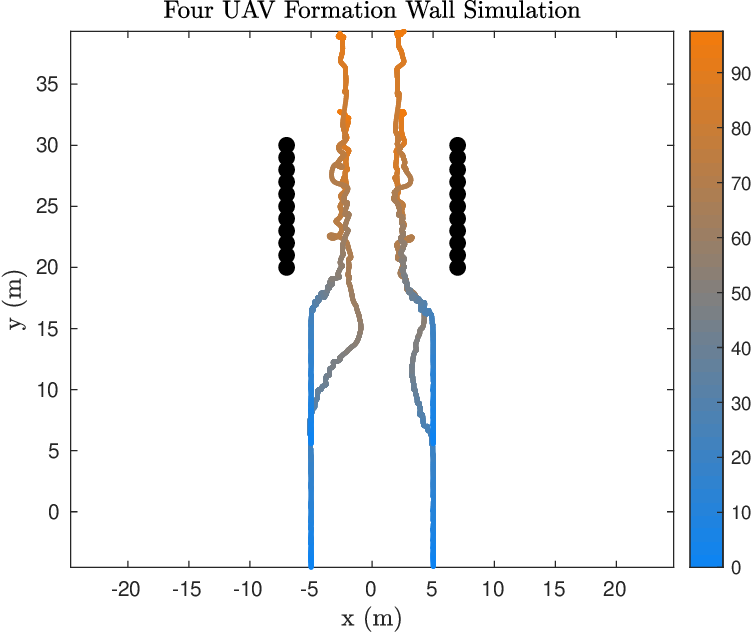}
    \caption{Robot trajectories for simulation of formation of four UAVs passing through a corridor. As the robots approach the corridor the formation starts to shrink until it reaches a stationary size that can pass through the corridor.}
    \label{fig:wall_sim}
\end{figure}

The presented method is implemented in the MRS UAV System simulator \cite{Baca2020TheVehicles}. The method is tested in the following scenario. A formation of drones fly towards a corridor. The expected behaviour is that the formation will start to shrink in order to pass through the corridor. If the corridor is sufficiently narrow the collision avoidance constraint will become active and the formation will cease shrinking, as can be seen in \cref{fig:wall_sim}.
\begin{figure}
    \centering
    \includegraphics[width=.75\linewidth]{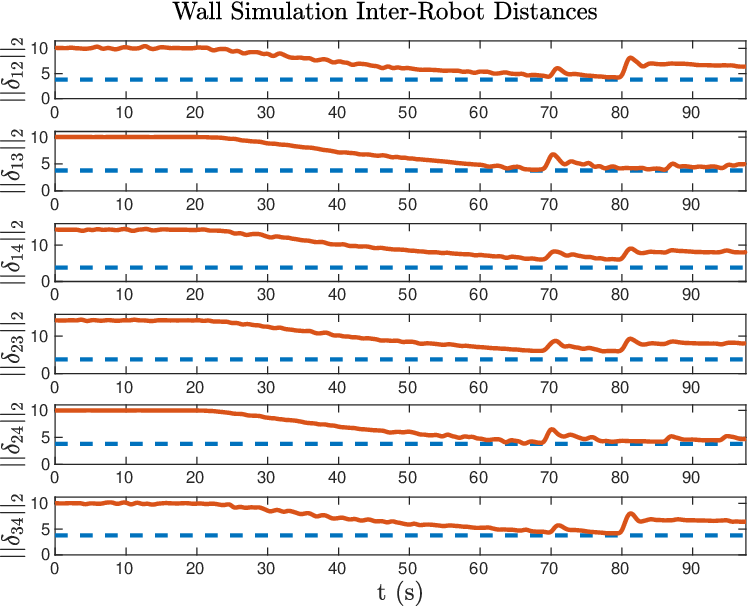}
    \caption{Inter-robot distances for simulation of four UAVs passing through a corridor. As the UAVs approach the corridor the inter-robot distances start to decrease until they reach the lower bound.}
    \label{fig:wall_sim_inter_robot_dist}
\end{figure}
In \cref{fig:wall_sim_inter_robot_dist} the inter-robot distances remain stationary until the robots have approached the corridor at which they start decreasing, due to the formation shrinking, until the constraint satisfaction step becomes active and ensures that the inter-robot distances remain greater than the lower bound in \cref{eq:ineq_dist}. Furthermore, it can be seen that as the inter-robot distances reach the lower bound the formation makes a sudden increase, which is due to the collision avoidance constraint, after which it starts shrinking again.
\section{Experiment Results}
The proposed method was tested in a real-world environment using MRS multi-rotor platforms \cite{Hert2023MRSEnvironments,Hert2023MRSSystems} equipped with the MRS UAV system~\cite{Baca2020TheVehicles} to provide state estimation, control, and reference tracking. The formation control method was deployed in a distributed manner, where each drone calculated its local parameter derivative and position reference based on information shared by other formation members. The UAVs also communicated with a base station, which provided formation changes commanded by the operator via a joystick through a WiFi channel.

\begin{figure}[ht]
    \centering
    \includegraphics[width=0.75\linewidth]{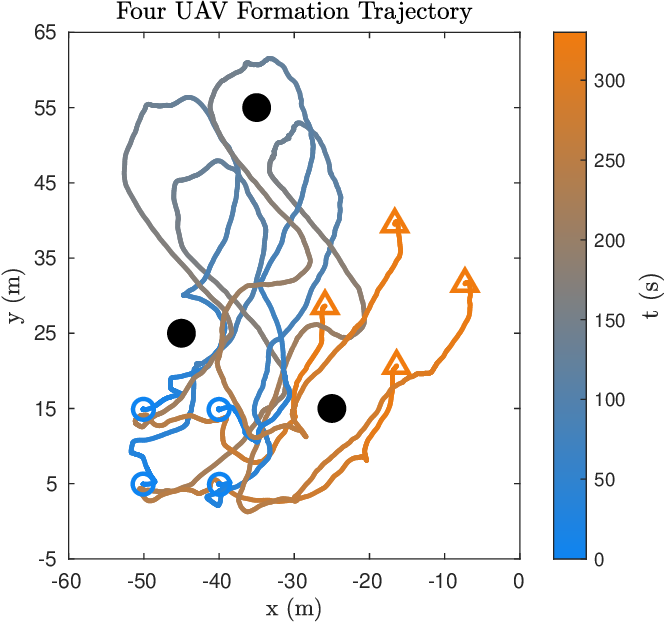}
    \caption{Full trajectory of UAV formation flying experiment with three cylindrical obstacles (black circles).}
    \label{fig:traj_full}
\end{figure}

\begin{figure}[ht]
    \centering
    \includegraphics[width=0.75\linewidth]{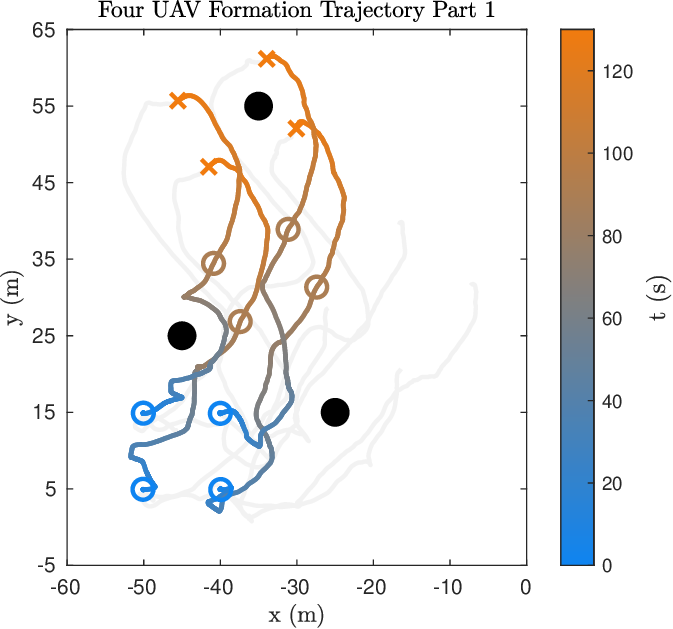}
    \caption{Trajectory from 0 seconds to 130 seconds of formation flying experiment.}
    \label{fig:traj_1}
\end{figure}

\begin{figure}[ht]
    \centering
    \includegraphics[width=0.75\linewidth]{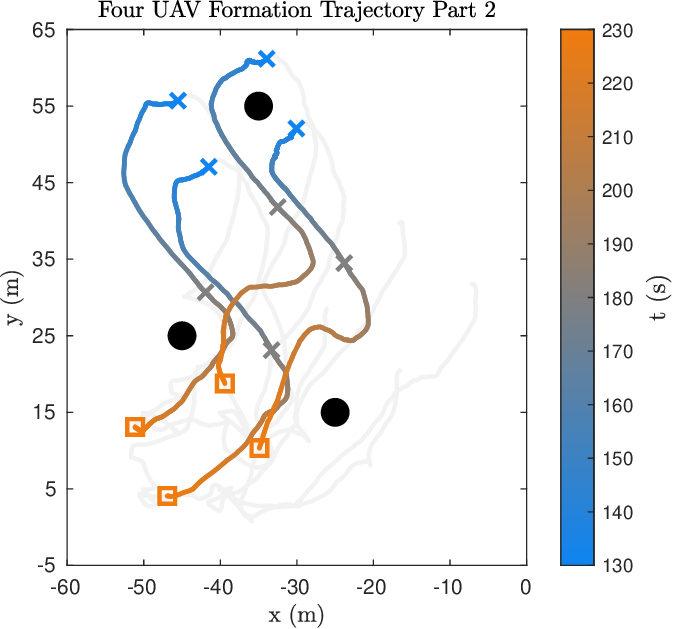}
    \caption{Trajectory from 130 seconds to 230 seconds of formation flying experiment.}
    \label{fig:traj_2}
\end{figure}

\begin{figure}[ht]
    \centering
    \includegraphics[width=0.75\linewidth]{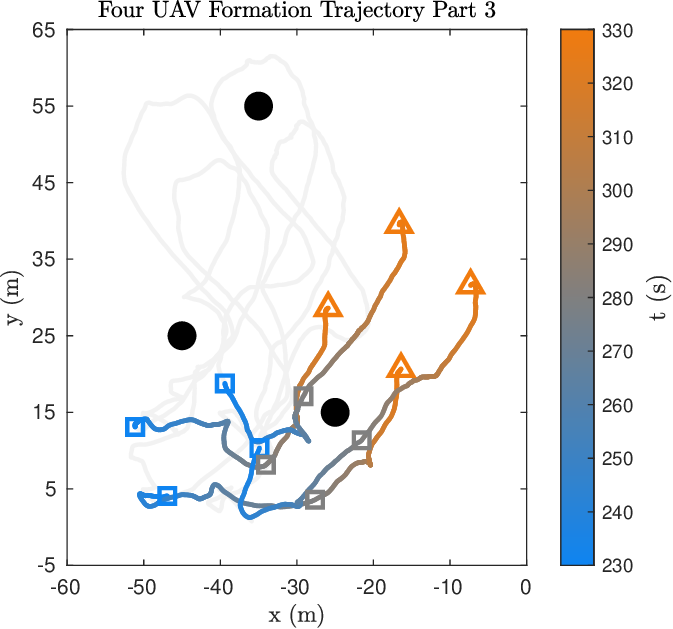}
    \caption{Trajectory from 230 seconds to 330 seconds of formation flying experiment.}
    \label{fig:traj_3}
\end{figure}

\begin{figure}[ht]
    \centering
    \includegraphics[width=0.75\linewidth]{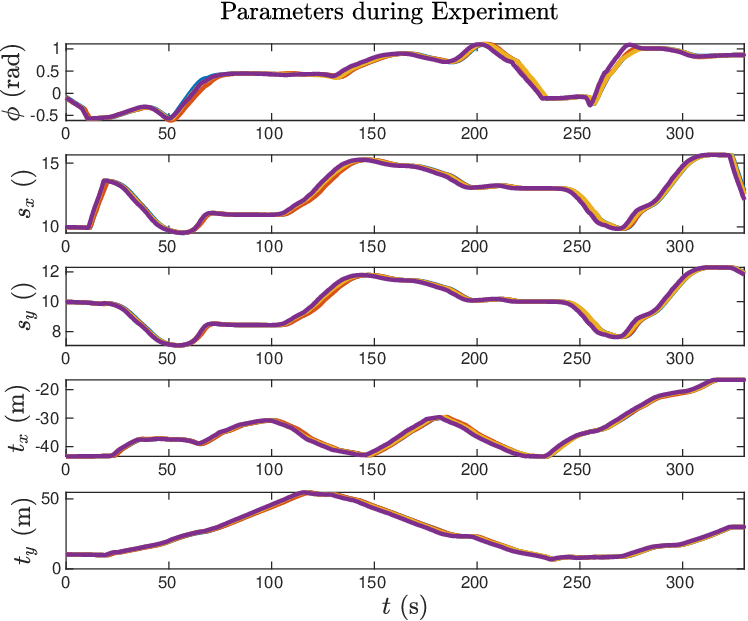}
    \caption{Formation parameters for UAV experiment.}
    \label{fig:param}
\end{figure}

\begin{figure}[ht]
    \centering
    \includegraphics[width=0.75\linewidth]{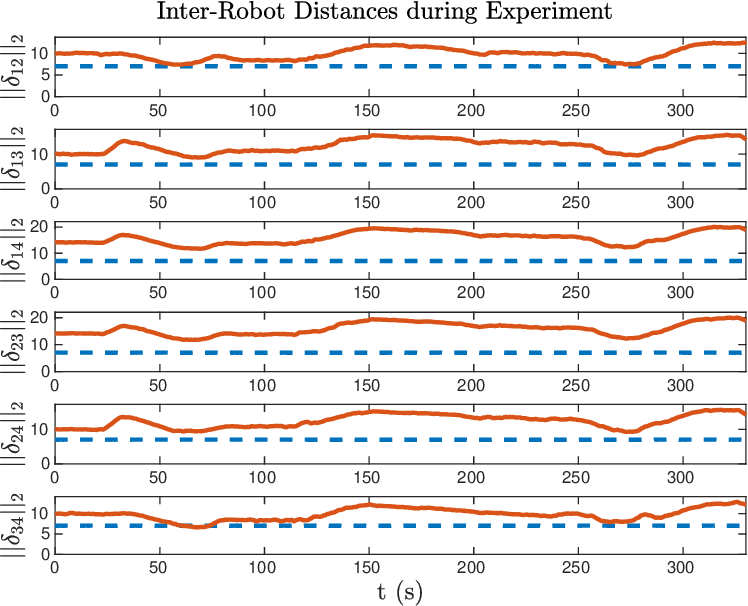}
    \caption{Inter-robot distances for experiment (orange solid) and lower bound (blue dashed).}
    \label{fig:exp_inter_robot_dist}
\end{figure}

The method was tested with four UAVs in a rectangular formation navigating around three cylindrical obstacles. The objective was to demonstrate that the operator could manipulate the formation without needing to consider obstacle positions. The full trajectory is shown in \cref{fig:traj_full}, with detailed segments in \cref{fig:traj_1,fig:traj_2,fig:traj_3}. The figures illustrate that the formation successfully deforms to navigate obstacles, without the operator manually adjusting the scale or rotation of the formation. As shown in \cref{fig:param}, the parameters remain mostly in consensus, with minor discrepancies due to obstacle avoidance influences. \cref{fig:exp_inter_robot_dist} displays the inter-robot distances relative to the minimum safe distance in \cref{eq:ineq_dist} needed to avoid collisions. The distances mostly stay above the lower bound, except for $||\bm{\delta_{34}}||_2$, which briefly dips below between 60 and 70 seconds as the formation navigates between the first two obstacles. This may result from parameter discrepancies or tracking errors, highlighting the importance of a clearance term. Distances also approach the lower bound around 270 seconds, coinciding with the robots navigating around the bottom-right obstacle.
\section{Discussion}
The following is a discussion of the presented method and the simulation and experiment results.
\subsection{Probabilistic Collision Avoidance}
In order to handle probabilistic collision avoidance in this paper, a two-step method was proposed. Firstly, due to there being no closed-form solution for calculating the probability of collision, an upper bounding approximation was proposed. This was then used with a parameterisation of the inter-robot distances to determine the safe set of scaling parameters that guarantee probabilistic collision avoidance. However, this has some limitations. In the process of determining the constraint set, the assumption was made that the discrepancy in the parameters was negligible. However, during the experiment results this was found not to be true. This, and possible the tracking error of the robots, lead the inter-robot distance for robot $3$ and $4$ to briefly dip below the lower bound while navigating around the two lower obstacles. The amount that the inter-robot distance dipped below the lower bound was not significant and was not detrimental to the robots as they still kept a safe distance, due to the desired clearance between the robots. For future research, it might be prudent to find a way to incorporate this discrepancy in the constraint, e.g., by adding an additional clearance which takes into account the discrepancy in the parameters.
\subsection{Damping and Soft Constraint}
As was seen in the corridor simulation, when the formation is continually shrunk until the collision avoidance constraint is activated, the formation enters into a limit cycle where it shrinks until the collision avoidance constraint is activated where it makes a sudden expansion, after which it starts shrinking again and repeats the process. Although this does not damage the robots, it is still an undesirable effect and could possibly be mitigated by adding a damping term to the parameter update. Another solution might be to add a soft constraint on the scaling parameters.
\subsection{Virtual Rigid Body}
In \cite{Mikkelsen2023DistributedConfiguration}, the parameter derivative was mapped back into the velocity space of the robots and applied with a position error feedback term. During the implementation of that method in the MRS UAV system \cite{Baca2020TheVehicles}, the rate that the nodes had to run at to be stable was unachievable within the ROS framework. Due to this, the method was applied to a VRB that the robots track using position controllers in order to decouple the dynamics of the UAVs from the formation, allowing the method to work at substantially lower rates.
\subsection{Obstacle Avoidance}
In this method, the obstacle avoidance APF is applied to the VRB and not to the robots directly. This might result in robots colliding with obstacles during transition to the desired formation. By limiting the velocity of the VRB to less than what the actual robots can achieve, it is assumed that any tracking error between the robots and the VRB is negligible. Even so, adding obstacle avoidance directly to the robots might be prudent.

\section{Conclusion}
A distributed method for having robots move in rigid formations, while ensuring an upper bound on the probability of collision by constraining the parameters of the formation, was presented. The method was tested in a manual teleoperation scenario where the desired change of the formation is controlled manually by the operator and the UAVs autonomously alter the formation to ensure collision avoidance with obstacles and each other. The method was demonstrated both in simulation and experiment.
\bibliography{references}

\end{document}